\documentclass[10pt,a4paper,twocolumn]{article}

\usepackage{graphicx} % Enhanced LaTeX Graphics

\usepackage{dcolumn}           % decimal-aligned tabular math columns
\newcolumntype{.}   {D{.}{.}{-1}} % column alignedd on the point separator '.'
\newcolumntype{d}[1]{D{.}{.}{#1}} % column centered on the point separator '.'
\newcolumntype{e}   {D{E}{E}{-1}} % column centered on the exponent 'E'
\newcolumntype{E}[1]{D{E}{E}{#1}} % column centered on the exponent 'E'

\usepackage{amsmath}
\usepackage{mathptmx}

\usepackage{amssymb}
\usepackage{amsthm}

\usepackage{float}

\usepackage{multicol}
\usepackage{subcaption}  % Required for subfigures
\usepackage{url}

\usepackage{breakurl}
\usepackage[breaklinks]{hyperref}
\usepackage{geometry}
\usepackage{fancyhdr}
\usepackage{booktabs}

\usepackage{setspace}

\usepackage{sectsty}

\newcommand{\myFontSize}{\fontsize{10}{12}\selectfont}
\renewcommand{\normalsize}{\fontsize{10}{12}\selectfont}
\sectionfont{\myFontSize}       % 10pt, Bold face (default)
\subsectionfont{\rm\myFontSize\itshape} % 10pt, Plain face

\usepackage{titlesec}
\titlespacing*{\section}{0pt}{10pt}{0pt}
\titlespacing*{\subsection}{0pt}{10pt}{0pt}

\usepackage{secdot}
\sectiondot{subsection}
\sectionpunct{section}{. }    % By default, \sectiondot places a \quad
\sectionpunct{subsection}{. } % after the number

\usepackage{caption}
\usepackage{xcolor} % Added package for colors
\usepackage{fancyhdr} % Added package for headers
\usepackage{lastpage}
\renewcommand{\headrulewidth}{0pt} % Remove the header rule line
\usepackage[numbers]{natbib}

\usepackage[shortcuts,acronym,nonumberlist]{glossaries}
\makeglossaries % generates the acronym list

\newacronym{LEO}{LEO}{low-Earth orbit}
\newacronym{PDE}{PDE}{partial differential equation}
\newacronym{SINDy}{SINDy}{sparse identification of nonlinear dynamics}
\newacronym{PINN}{PINN}{physically-informed neural network}
\newacronym{FROLS}{FROLS}{Forward Regression Orthogonal Least-Squares}
\newacronym{RSO}{RSO}{Resident Space Object}
\newacronym{SVM}{SVM}{Support Vector Machine}
\newacronym{SSO}{SSO}{Sun-synchronous Orbit}

\begin{document}

% Begin one column section for title and abstract
%
% http://www.faqs.org/faqs/de-tex-faq/part5/
\twocolumn[
\begin{@twocolumnfalse}

\vspace{0pt}
\begin{center}
    % \selectfont\fontsize{10}{0}\selectfont IAC-23,A6,IPB,5,x76700
    
    \vspace{15pt}
    \textbf{Finding Real-World Orbital Motion Laws from Data}
    
    \vspace{10pt}
    \textbf{
        \selectfont\fontsize{10}{0}\selectfont João~Funenga~\textsuperscript{a*},~Marta~Guimarães\textsuperscript{b},~Henrique~Costa\textsuperscript{b},~Cláudia~Soares~\textsuperscript{a},
    }
\end{center}

% \vspace{-30pt}
% \maketitle

\vspace{-10pt} 
\begin{flushleft}
\textsuperscript{a}\textit{
        \fontfamily{ptm}\selectfont\fontsize{10}{12}\selectfont FCT-UNL, Portugal},
        \underline{j.funenga@campus.fct.unl.pt}, 
        \underline{claudia.soares@fct.unl.pt}
    \\
    \textsuperscript{b}\textit{
    \fontfamily{ptm}\selectfont\fontsize{10}{12}\selectfont Neuraspace, Portugal}, \underline{marta.guimaraes@neuraspace.com}
    \\
    
    \textsuperscript{*}\fontfamily{ptm}\selectfont\fontsize{10}{12}\selectfont Corresponding Author  
\end{flushleft}

%%%%%%%%%%%%%%%%%%%%%%%%%%%%%%%%%%%%%%%%%%%%%%%%%%%%%%%%%%%%%%%%%%%%%%
% ABSTRACT & KEYWORDS
%%%%%%%%%%%%%%%%%%%%%%%%%%%%%%%%%%%%%%%%%%%%%%%%%%%%%%%%%%%%%%%%%%%%%%
% \vspace{10pt} 
\begin{abstract}
A novel approach is presented for discovering~\glspl{PDE} that govern the motion of satellites in space. The method is based on~\gls{SINDy}, a data-driven technique capable of identifying the underlying dynamics of complex physical systems from time series data.~\gls{SINDy} is utilized to uncover~\glspl{PDE} that describe the laws of physics in space, which are non-deterministic and influenced by various factors such as drag or the reference area (related to the attitude of the satellite).
In contrast to prior works, the physically interpretable coordinate system is maintained, and no dimensionality reduction technique is applied to the data. By training the model with multiple representative trajectories of \gls{LEO} – encompassing various inclinations, eccentricities, and altitudes – and testing it with unseen orbital motion patterns, a mean error of around 140 km for the positions and 0.12 km/s for the velocities is achieved. The method offers the advantage of delivering interpretable, accurate, and complex models of orbital motion that can be employed for propagation or as inputs to predictive models for other variables of interest, such as atmospheric drag or the probability of collision in an encounter with a spacecraft or space objects.
In conclusion, the work demonstrates the promising potential of using~\gls{SINDy} to discover the equations governing the behaviour of satellites in space. The technique has been successfully applied to uncover~\glspl{PDE} describing the motion of satellites in~\gls{LEO} with high accuracy. The method possesses several advantages over traditional models, including the ability to provide physically interpretable, accurate, and complex models of orbital motion derived from high-entropy datasets. These models can be utilised for propagation or as inputs to predictive models for other variables of interest.

%%
%% Keywords (max 5)
%%

\end{abstract}

% End one column section (begin default two columns)
\end{@twocolumnfalse}
]

% \section*{Acronyms/Abbreviations}
% % \glsnogroupskiptrue
% \printglossary[type=acronym,title={},style=long,nogroupskip=true]
% \setcounter{table}{0}

\section{Introduction}

The successful operation of satellites for communication, scientific research, and navigation relies on precisely determining their state vectors, i.e., the position and velocity over time. This information is crucial in optimising propulsion system planning, minimising fuel consumption, and maximising operational efficiency.

Satellites are subjected to various influential forces, including drag, gravitational forces from celestial bodies, and solar radiation pressure~\cite{srpeffect}. Accurately determining the state vectors provides a comprehensive understanding of these forces and their impact on the trajectory of satellites. By considering these factors holistically, satellite operators can anticipate and mitigate their effects, ensuring the satellite stays on its intended path. Failure to adequately consider these factors can result in substantial deviations from the nominal orbit, potentially spanning several hundred-kilo meters~\cite{dragError}. Such errors might have significant consequences, compromising communication links, hindering scientific observations, and impeding navigation systems. Thus, a precise prediction of the state of a satellite, utilising orbital motion laws, enables proactive manoeuvre planning and necessary corrections, enhancing mission success and overall satellite performance.

Furthermore, as the demand for reliable and accurate satellite services continues to rise, accompanied by an increase in the number of satellites launched into space, meticulous planning and precise position prediction become even more critical~\cite{increasesatellites}. Maintaining an organised and controlled satellite environment is essential to prevent collisions and satellite interference. By precisely determining the state vectors, satellite operators can efficiently optimise satellite deployments, plan collision avoidance manoeuvres, and allocate orbital slots.

However, accurately modelling trajectory-altering effects poses significant challenges. One notable example is atmospheric density, which directly impacts the motion of a satellite. Predicting atmospheric density is difficult due to its dependence on various factors with high stochasticity, including solar activity, temperature variations, and atmospheric composition~\cite{difficultypredictdensity}. The interactions between these elements introduce uncertainties into the calculations, making it difficult to precisely account for atmospheric drag and its effect on the trajectories of satellites.

The atmospheric density variability highlights the challenges in modelling such trajectory-altering effects. Even subtle changes in atmospheric conditions can result in significant variations in drag forces experienced by a satellite, leading to perturbations affecting the planned path.

\subsection{Related Work}

In recent years, there has been growing interest in applying the~\gls{SINDy} algorithm to extract governing equations and uncover hidden dynamics from observational data. While~\gls{SINDy} has shown promise in various applications, such as discovering equations of motion and identifying relevant terms in dynamical systems, its specific utilisation for predicting state vectors of satellites has received limited attention in the literature.
  
Several studies have successfully applied \gls{SINDy} to various domains, such as fluid dynamics~\cite{fluiddynamics} or biochemical systems~\cite{biochemical}.
Moreover, \gls{SINDy} has proven effective in identifying the equations of motion and understanding system behaviour in mechanical systems. For example, in~\cite{pendulum}, the authors employed~\gls{SINDy} to reveal the underlying mathematical model governing the motion of a damped double pendulum. The authors successfully captured the system dynamics and accurately predicted its motion using the derived equations by analysing experimental data.
While \gls{SINDy} has not been applied explicitly for predicting state vectors of satellites, it has indeed found applications within space research. One such notable example is~\cite{autoencodersindy}, where \gls{SINDy} was employed to derive best-fitting differential equations governing the spatial and temporal evolution of the thermospheric density field. This approach allowed for real-time density estimation, an essential factor in understanding the dynamics of space objects in \gls{LEO} due to atmospheric drag.
In~\cite{autoencodersindy}, the ability of \gls{SINDy} to extract governing equations from observed data while promoting sparsity was demonstrated in the space domain, enabling a concise and interpretable representation of the thermospheric density dynamics. The effectiveness of the method in this context showcases its potential for understanding complex systems within space research.

The rapid expansion of global satellite communication companies, advancements in miniaturised satellites, and revolutionary ideas such as autonomous nanosatellite swarms~\cite{nanoswarm} have significantly amplified the potential for conflicts and collisions among these orbiting entities. Consequently, ensuring accurate and timely trajectory predictions for space objects has become crucial to establish a solid foundation for present and future space situational awareness systems. The traditional physics-based models used for orbit prediction often fail to achieve the required accuracy, leading to collisions due to the lack of essential information about the space environment and characteristics of the~\glspl{RSO}, which can be challenging to acquire.
Machine Learning techniques have been used to predict satellite state vectors. In~\cite{svmStateVectors}, the authors address the challenges of efficiently and accurately predicting the orbit of~\glspl{RSO} for space situational awareness and collision avoidance purposes.
The growing population of space objects in orbits in \gls{LEO} has recently become a primary concern for space situational awareness~\cite{increasedebris}. 
To overcome these limitations, the authors in~\cite{svmStateVectors} hypothesise that a machine learning approach can learn the underlying patterns of orbit prediction errors from historical data. They specifically explore using~\glspl{SVM}~\cite{svm} to enhance the accuracy of orbit predictions. The \gls{SVM} model is designed and trained at a current epoch and then utilised to reduce the orbit prediction error at a future epoch.
Through simulations involving~\glspl{RSO} in a~\gls{SSO}, the study demonstrates that the trained \gls{SVM} model effectively captures the underlying relationships between the learning variables and provides desirable predictions for the orbital motion. It shows promising results with good average and individual performance in reducing prediction errors.
The paper also indicates that there is a limit to the improvements once sufficient data have been utilised for training the model, and one drawback is that it needs to be updated frequently in practice. Orbit predictions should not be made too far into the future.

The findings of this paper add to the body of research exploring machine learning approaches for orbit prediction accuracy improvement. While our work focuses on applying SINDy to discover the governing equations of satellite motion, using an \gls{SVM} in orbit prediction showcases the potential of various machine learning techniques in enhancing space-related predictions. The combination of diverse approaches can contribute to advancing space situational awareness and managing space objects in the future.
However, despite the wide-ranging applications of~\gls{SINDy}, its direct application for predicting state vectors of satellites from observational data remains largely unexplored.
To the best of our knowledge, no previous work has specifically investigated the use of \gls{SINDy} or similar methods to directly predict state vectors of satellites, maintaining the physical meaning of the variables predicted.
This represents a significant gap in the literature, as an accurate and interpretable prediction of satellite trajectories is crucial for multiple space-related applications, including orbit determination, collision avoidance, and mission planning.
Through this research, we aim to demonstrate the effectiveness of~\gls{SINDy} for predicting state vectors and contribute to the broader field of satellite trajectory analysis. By leveraging the vast amount of available observational data in conjunction with high-fidelity simulators, we strive to enhance our understanding of the dynamics and interactions that drive satellite motion, thus enabling improved satellite operations.

\subsection{Contributions}

In this work, we demonstrate the potential of using the~\gls{SINDy} algorithm to model and uncover a~\glspl{PDE} that describes well in expectation the non-deterministic laws of physics in space, accounting for various influential factors such as drag and the reference area of the surface normal to the satellite motion. 

Unlike previous work that applied dimensionality reduction techniques on data~\cite{hasdmstateoftheart}, using this data-driven technique, the physical system retains its coordinate system and thus is interpretable. Indeed, such representation allows for a clear understanding of the underlying physics involved.

By training the model with multiple representative trajectories of~\gls{LEO} objects, the proposed solution achieves high-level accuracy in predicting both the position and the velocity, contributing to advancing satellite trajectory prediction and enhancing the understanding of the dynamics of the space environment and satellite operations.

\section{SINDy: Nonlinear Governing Equations from Data}

\gls{SINDy}, proposed in~\citet{sindy}, aims to extract governing equations from observed data while prioritizing sparsity. The fundamental assumption underlying \gls{SINDy} is that the discovered equations will consist of only a few terms. This sparsity assumption enhances the robustness of the model by reducing sensitivity to noise and preventing the identification of extra residual terms solely due to noisy input. Such an approach introduces a trade-off between complex and sparse models. On the one hand, a complex model accurately captures the intricacies of a system, but it risks overfitting the specific dataset used for model discovery.
On the other hand, a sparse model is less complex, incorporates fewer terms, and avoids overfitting. However, it may sacrifice some accuracy compared to the more complex model. In this context, model complexity refers to the number of terms in the discovered equations.
We will use a framework developed in Python to leverage the capabilities of \gls{SINDy}~\cite{pysindy}.

Considering a set of measurements $x(t)\in \mathbb{R}^{n}$ at different points in time $t$, ~\gls{SINDy} models the time evolution of such measurements in terms of a nonlinear function $f$.
Thus, the dynamical system for $x(t)$ is given by
\begin{equation}
    \begin{aligned}
    \label{eqn:sindyeq}
        \frac{d}{dt}x(t) = f(x(t)),
    \end{aligned}
\end{equation}
where $x\left( t \right)=\left[x_1(t), x_2(t), \cdots , x_n(t)  \right]^{T}$ represents the state of the physical system at time $t$, and $f(x(t))$ constrains how the system evolves over time.

The implementation of \gls{SINDy} requires a dataset comprising measurements collected at specific time instances, $t_1, t_2, \cdots, t_n$. Furthermore, the corresponding time derivatives of such measurements are also needed. These datasets are then organised into two matrices: $X$, containing the measurements, and $\dot{X}$, which stores the corresponding derivatives. The user also provides a library of candidate functions, $\theta (X)$. Such a library consists of a set of basis functions that will be applied to the data. For example, the polynomial library would be defined as
\[
\Theta(X) = 
\left[
  \begin{array}{cccccc}
     \vert & \vert & \vert & \vert & \vert & \\
    1   & X & X^{P_2} & X^{P_3} & X^{P_4} &\cdots   \\
    \vert & \vert & \vert & \vert & \vert &  
  \end{array}
\right]
\]
where the mth-degree Vandermonde matrix (of polynomials up to degree m) is given by
\begin{figure}[htbp]
\centering
\begin{tabular}{cc}
\begin{minipage}{0.9\linewidth}
\resizebox{\linewidth}{!}{$
X^{P_m} = \begin{bmatrix} 
    x^2_1(t_1) & x_1(t_1)x_2(t_1) & \dots & x^2_2(t_1) & \cdots & x^m_n(t_1) \\
    x^2_1(t_2) & x_1(t_2)x_2(t_2) & \dots & x^2_2(t_2) & \cdots & x^m_n(t_2) \\
    \vdots     &       \vdots     &  \ddots &  \vdots   & \ddots  & \vdots \\
    x^2_1(t_m) & x_1(t_m)x_2(t_m) & \dots & x^2_2(t_m) & \cdots & x^m_n(t_m) \\
    \end{bmatrix}.
$}
\end{minipage}
&
\begin{minipage}{0.5\linewidth}
% Second column content
\end{minipage}
\end{tabular}
\label{fig:resizedmatrix}
\end{figure}

We want to find a set of sparse coefficient vectors
\[
\Xi(X) = 
\left[
  \begin{array}{cccc}
    \vert & \vert &  & \vert\\
    \xi_1 & \xi_2 & ... & \xi_n   \\
    \vert & \vert &  & \vert
  \end{array}
\right] ,
\]
where $\xi_i$ defines the coefficients for a linear combination of the basis functions from $\Theta (X)$. Thus, the approximation problem underlying \gls{SINDy} can be defined as
\begin{equation}
    \label{eq:approximation_problem}
    \dot{X} = \Theta (X)\Xi.
\end{equation}
In practical scenarios, it is common for the data matrices $X$ and $\dot{X}$ to be affected by noise, resulting in deviations from the nominal identity in~\eqref{eq:approximation_problem}. In cases where the measurements in $X$ are relatively clean but the derivatives in $\dot{X}$ are noisy, the equation can be modified to account for this noise
\begin{equation}
    \begin{aligned}
    \label{eq:xdotmatrix}
        \dot{X} = \Theta (X)\Xi + \eta Z , 
    \end{aligned}
\end{equation}
where $Z$ corresponds to a matrix of i.i.d.\ Gaussian random variables with mean zero and standard deviation $\eta$.

\section{Data Sources}  \label{sec:data}

Due to the scarcity of measurements from real-world data per trajectory, we employed a realistic high-fidelity propagator used in Neuraspace that accounts for various exogenous perturbations, including solar radiation pressure, atmosphere density models, and gravity variations due to the oblateness of the Earth. Additionally, we considered internal information specific to each satellite, 
such as its reference area, drag coefficient, and mass. Utilising this propagator, we generated a more finely-grained dataset, significantly increasing the number of measurements available for analysis. This data augmentation process enables the application of the \gls{SINDy} methodology.

The resultant dataset consists of state vectors, each representing the complete state of a satellite. These vectors encompass both the position and velocity values along each axis, providing a comprehensive depiction of the motion of the satellite in three-dimensional space.
With this enriched dataset, we can explore the capabilities of SINDy in uncovering the underlying governing equations of the dynamics of the satellite system.

\section{Candidate Nonlinear Functions}

~\gls{SINDy} requires the appropriate choice of a coordinate system and function basis to capture the sparse dynamics of the system accurately. However, such steps can be challenging and nontrivial~\cite{sindy}. In this context, domain-specific knowledge of the underlying physics can be priceless. By leveraging physics knowledge, one can use the power of data to guide the selection of appropriate coordinates and simplify the dynamical model of the system. This interplay between domain expertise and data-driven analysis facilitates the discovery of meaningful and interpretable system behaviour models.

In this work, two different function bases were considered to explore the modelling of the nonlinear dynamics of the system. The first function basis is grounded on the underlying physics of the problem, aiming to capture the intrinsic relationships and principles governing the behaviour of the system. The second function basis consists of polynomial functions, which are more general and widely applicable, enabling the exploration of more straightforward and more interpretable representations of the data.

\subsection{Domain-Driven Custom Functions}

We begin by utilising a custom functions library, with the primary objective of assessing the capability of \gls{SINDy} to identify the correct terms among the available options for constructing the equations. As mentioned in Section~\ref{sec:data}, the data consists of a time series of state vectors containing the positions and velocities of a given object. Thus, we seek to find the first-order~\glspl{PDE} of the underlying system. For a given state vector,
$$
    w = 
    \begin{bmatrix}
        x & y & z & \Dot{x} & \Dot{y} & \Dot{z}
    \end{bmatrix},
    $$
to find equations that accurately describe 
\begin{equation}
    \label{wdot}
    \dot{w} = 
        \begin{bmatrix}
            \Dot{x} & \Dot{y} & \Dot{z} & \Ddot{x} & \Ddot{y} & \Ddot{z}
        \end{bmatrix}
    .
\end{equation}

From the orbital motion equations~\cite{mechbook}, we can approximate $\dot{w}$ by neglecting the contribution of external forces. Thus, $\dot{w}$ is given by
\begin{equation}
    \label{wdotcalculated}
    \dot{w} \approx 
    \begin{bmatrix}
        \Dot{x} & \Dot{y} & \Dot{z} & \gamma x & \gamma y & \gamma z
    \end{bmatrix}
\end{equation}
where $ \gamma = \frac{\mu}{(x^2 + y^2 + z^2)^{3/2}}$, and $\mu$ is the standard gravitational parameter of the Earth.
In this sense, the proposed domain-driven custom library contains the terms in~\eqref{wdotcalculated}, and the main purpose of this trial is to study how can \gls{SINDy} recover the well-known parameters for the standard solution of orbital motion.

\subsection{Polynomial Functions}
As mentioned, a polynomial library was also considered. The goal of incorporating polynomial functions is to explore a more general and flexible approach to represent the observed state vectors. By considering polynomial terms of varying degrees, we aim to assess the possibility of capturing the nonlinear system dynamics using simpler and more interpretable terms without relying on domain-specific knowledge. In this work, terms up to degree four were considered, which allow for capturing a wide range of nonlinear relationships and interactions within the system. By limiting the degree to four, we aim for a practical balance, enabling us to capture important nonlinearities while maintaining a manageable number of terms in the model.

\section{Results}
To assess the performance of our approach, i.e., the effectiveness of the equations obtained through the application of \gls{SINDy}, such equations were used to propagate the data over time. The resulting data points were then compared to the observed trajectory of the satellite. This comparative analysis enabled us to evaluate the predictive accuracy of the methodology and its reliability.

\subsection{Choice of the SINDy Optimizer}  

One of the key decisions when applying \gls{SINDy} is the choice of the model optimiser~\cite{variancesindy}. 
However, after choosing one, it is still highly dependent on multiple factors revolving around the data fed (if they are standardised or not, if multiple trajectories are used, if the data contain drag or not, and finally if they are the result of custom functions or polynomial terms), the differentiation method chosen and even with everything fixed, there is still some variance on the identified equations comparing multiple runs~\cite{variancesindy, variancesindy2}.
Several optimization methods are used in the literature~\cite{pysindy}.

However, we have decided to proceed with the \gls{FROLS}~\cite{frolsoptimizer} since it offers advantages in terms of interpretability, computational efficiency, and robustness to noise. \gls{FROLS} tries to solve the following optimisation problem:

\begin{equation}
\begin{aligned}
\label{mineq}
\min_{v} \quad & \|t - Av\|_2^2 + \alpha\|v\|_2^2 + b \|v\|_0
\end{aligned}
\end{equation}
where $b = \kappa N$, $N$ is the condition number of the matrix $\theta$ which corresponds to the function library whose columns represent the set of basis functions and $\|u\|_0 = \sum^N_ {n=1} |u_n|^0$ corresponding to the L0 norm that is the number of non-zero entries in $u$.
This optimiser has two tunable parameters, $\alpha$ and $\kappa$. $\alpha$ represents the optional L2 regularisation on the weight vector to enforce smaller coefficients and $\kappa$ is also an optional parameter that if used, computes the mean squared error with an extra L0 regularisation term with strength equal to $b$ above-mentioned.

\subsection{Definition of the Domain-Driven Custom Library}

\begin{figure*}[t]
    \centering
    \begin{subfigure}[b]{0.3\linewidth}
        \includegraphics[width=\linewidth]{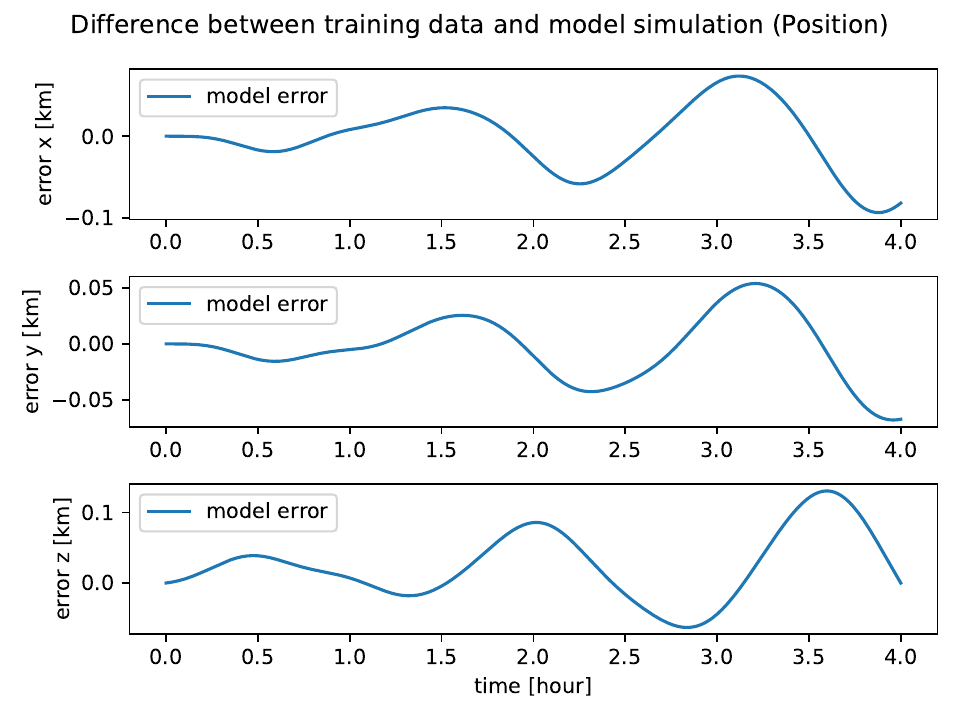}
        \caption{Errors in the position.}
        \label{fig:sub1}
    \end{subfigure}
    \hfill
    \begin{subfigure}[b]{0.3\linewidth}
        \includegraphics[width=\linewidth]{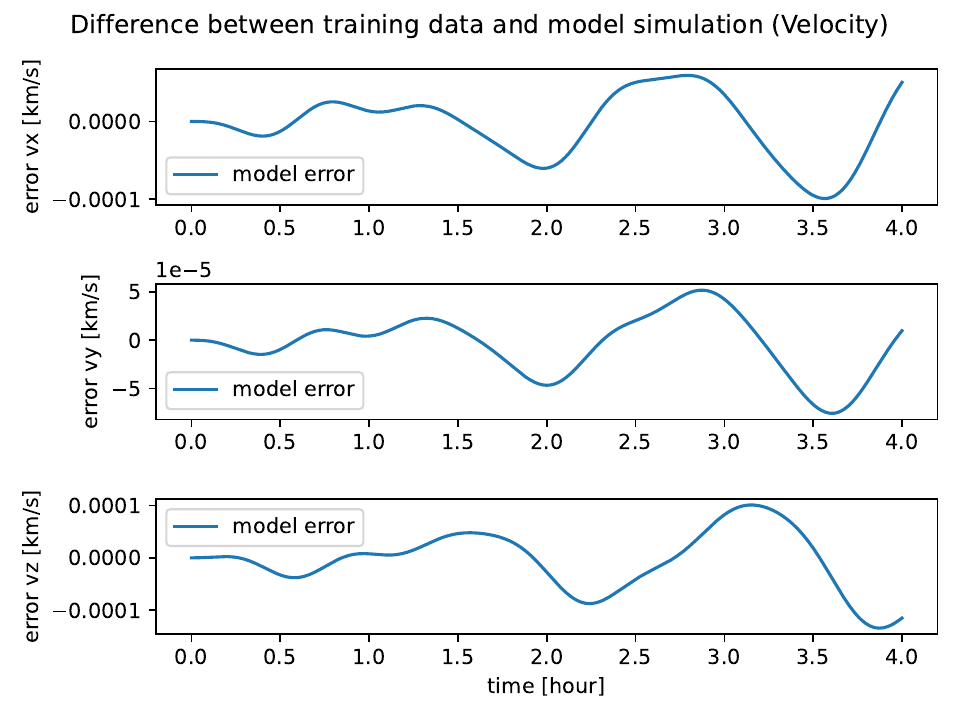}
        \caption{Errors in the velocity.}
        \label{fig:sub2}
    \end{subfigure}
    \hfill
    \begin{subfigure}[b]{0.3\linewidth}
        \includegraphics[width=\linewidth]{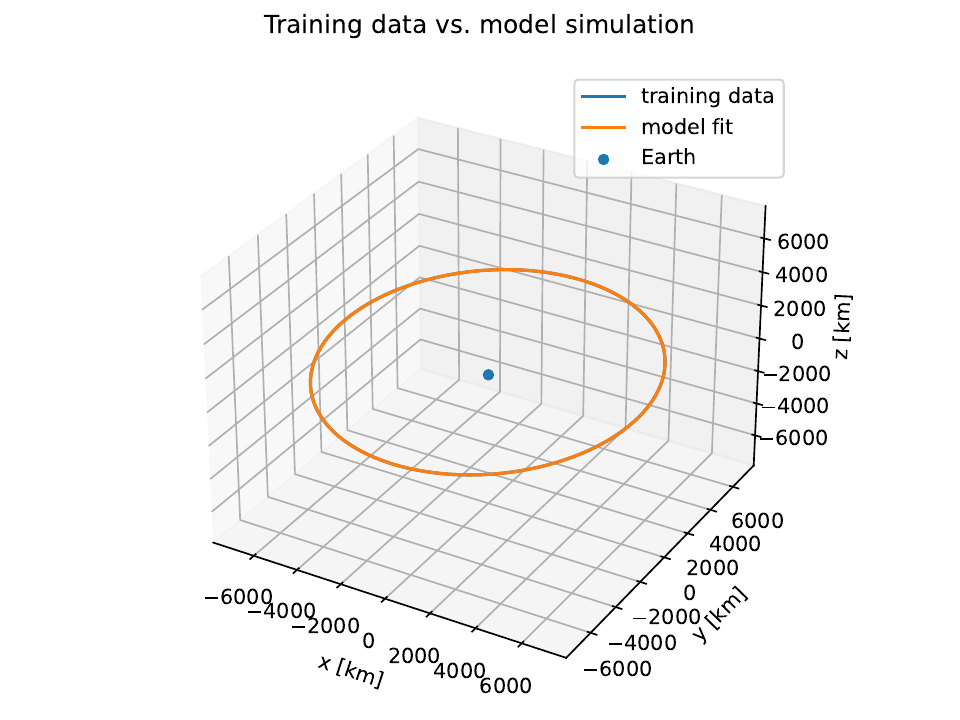}
        \caption{Orbital trajectory.}
        \label{fig:sub3}
    \end{subfigure}
    \caption[Single Orbit Performance with Custom Functions]{Errors obtained when comparing the true values with the ones obtained using the custom library.}
    \label{fig:singorbcust}
\end{figure*}

For numerical stability purposes, we used the data in kilometre units.
After applying \gls{SINDy} to the data obtained from propagating the initial state vector over three hours, the solution, as represented in~\eqref{wdotcalculated}, was accurately determined. Notably, the gravitational parameter was correctly identified up to the ninth decimal place, indicating the precision achieved in the estimation process.
\begin{align*}
    \text{Real Value} &= 3.986004418 \times 10^{14} m^3 s^{-2} \\
    \text{Predicted Value} &= 3.986004412 \times 10^{14} m^3 s^{-2}
\end{align*}

When looking at the errors attained for a span of four hours as seen in~\ref{fig:singorbcust}, it is possible to see that the errors are low, having mean values for the positions as -0.0022~km, -0.0033~km and 0.025~km for each axis respectively and for the velocities $-5.63 \times 10^{-6}$,$-4.66 \times 10^{-6}$ and $-4.38 \times 10^{-6}$ for each axis.
However, one has to keep in mind that here we have no external non-conservative forces acting on the satellite that would make the predictions harder. This experiment has the objective of finding the standard gravitational parameter of the Earth from data. It is also important to state that we are specifically saying which terms will have to be present in the identified \glspl{PDE} which eases the difficulty of the predictions of the model.

\subsection{Polynomial Library}
When applying the polynomial library, \gls{SINDy} successfully identified the first-order \glspl{PDE} governing the positions of the system accurately. However, it did not yield any suitable equations to describe the dynamics of the velocities within the given dataset, as illustrated below.

\begin{multicols}{2}
  \noindent
  \begin{minipage}{\columnwidth}
  {\small
    \begin{align*}
      \Dot{x}_{\text{calculated}} &= c_1 \Dot{x}\\
      \Dot{y}_{\text{calculated}} &= c_2 \Dot{y}\\
      \Dot{z}_{\text{calculated}} &= c_3 \Dot{z}
    \end{align*}
    }
  \end{minipage}
  \vspace{10pt} % Adjust the value as needed
  \noindent
  \begin{minipage}{\columnwidth}
  {\small
    \begin{align*}
      \Ddot{x}_{\text{calculated}} &= 0.00\\
      \Ddot{y}_{\text{calculated}} &= 0.00\\
      \Ddot{z}_{\text{calculated}} &= 0.00
    \end{align*}
    }
  \end{minipage}
\end{multicols}
\noindent
where $c_1 = c_2 = c_3 = 1.00$.

Upon examining the range of values for each variable, we observed significant differences in orders of magnitude. This discrepancy posed a challenge for the optimisation step of \gls{SINDy}, which aims to identify the correct \glspl{PDE} that best fits the data. To address this issue, we considered standardising the data to alleviate the impact of varying scales.

By fitting the model with standardised data, we achieved a straightforward polynomial \glspl{PDE} that accurately described the orbit:

\begin{multicols}{2}
  \noindent
  \begin{minipage}{\columnwidth}
  {\small
    \begin{align*}
      \dot{x}_{\text{calculated}} &= c_1 \dot{x}\\
      \dot{y}_{\text{calculated}} &= c_2 \dot{y}\\
      \dot{z}_{\text{calculated}} &= c_3 \dot{z}
    \end{align*}
  }
  \end{minipage}
  
  \vspace{10pt} % Adjust the value as needed
  
  \noindent
   \begin{minipage}{\columnwidth}
   {\small
    \begin{align*}
      \ddot{x}_{\text{calculated}} &= c_4 x\\
      \ddot{y}_{\text{calculated}} &= c_5 y\\
      \ddot{z}_{\text{calculated}} &= c_6 z
    \end{align*}
   }
   \end{minipage}
\end{multicols}

%\vspace{20px}
%
\noindent
where $c_1 = c_2 = c_3 = 0.001$ and $c_4 = c_5 = c_6 = -0.001$.

\begin{figure*}[t]
    \centering
    \begin{subfigure}[b]{0.3\linewidth}
        \includegraphics[width=\linewidth]{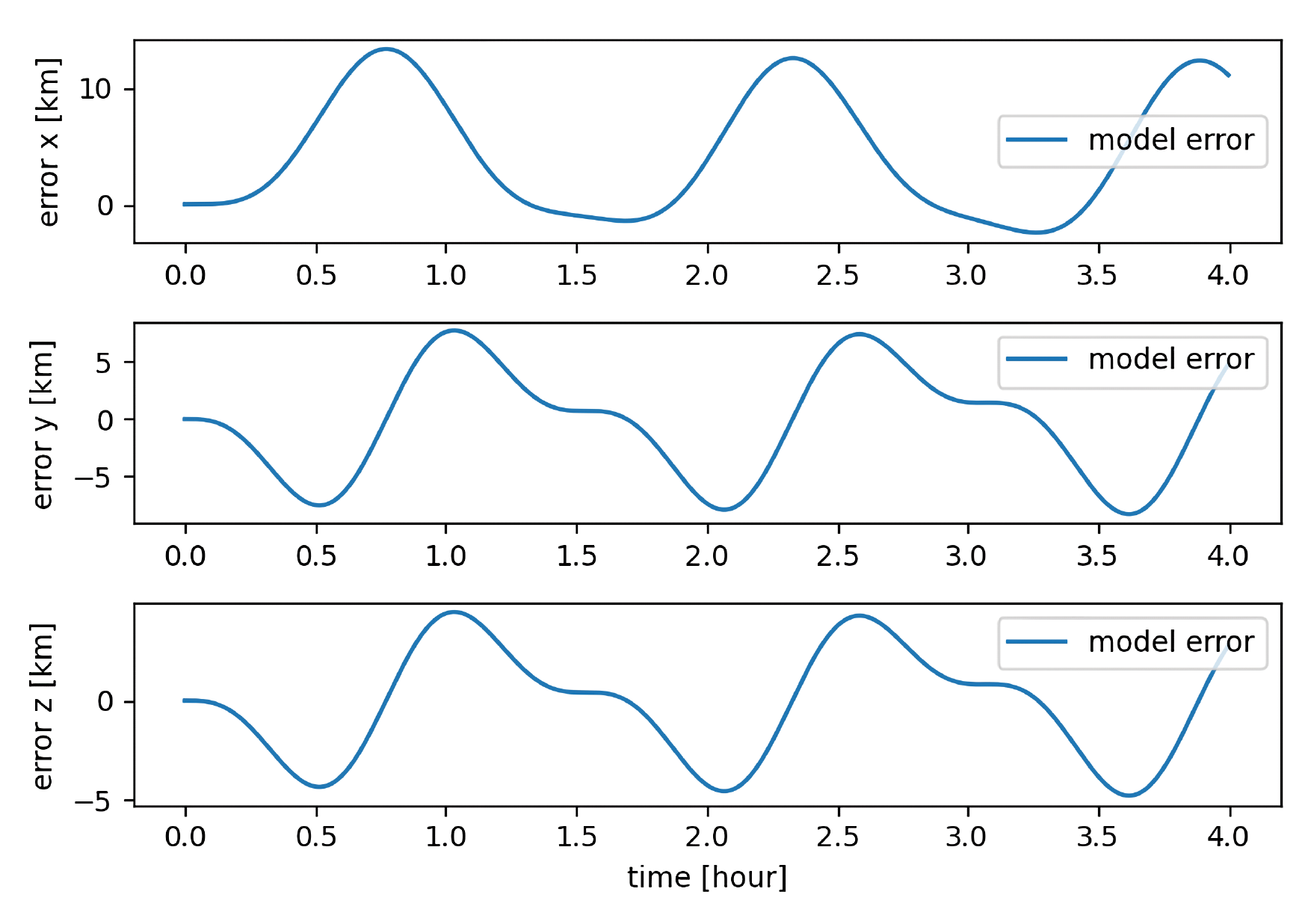}
        \caption{Errors in the position.}
        \label{fig:sub1}
    \end{subfigure}
    \hfill
    \begin{subfigure}[b]{0.3\linewidth}
        \includegraphics[width=\linewidth]{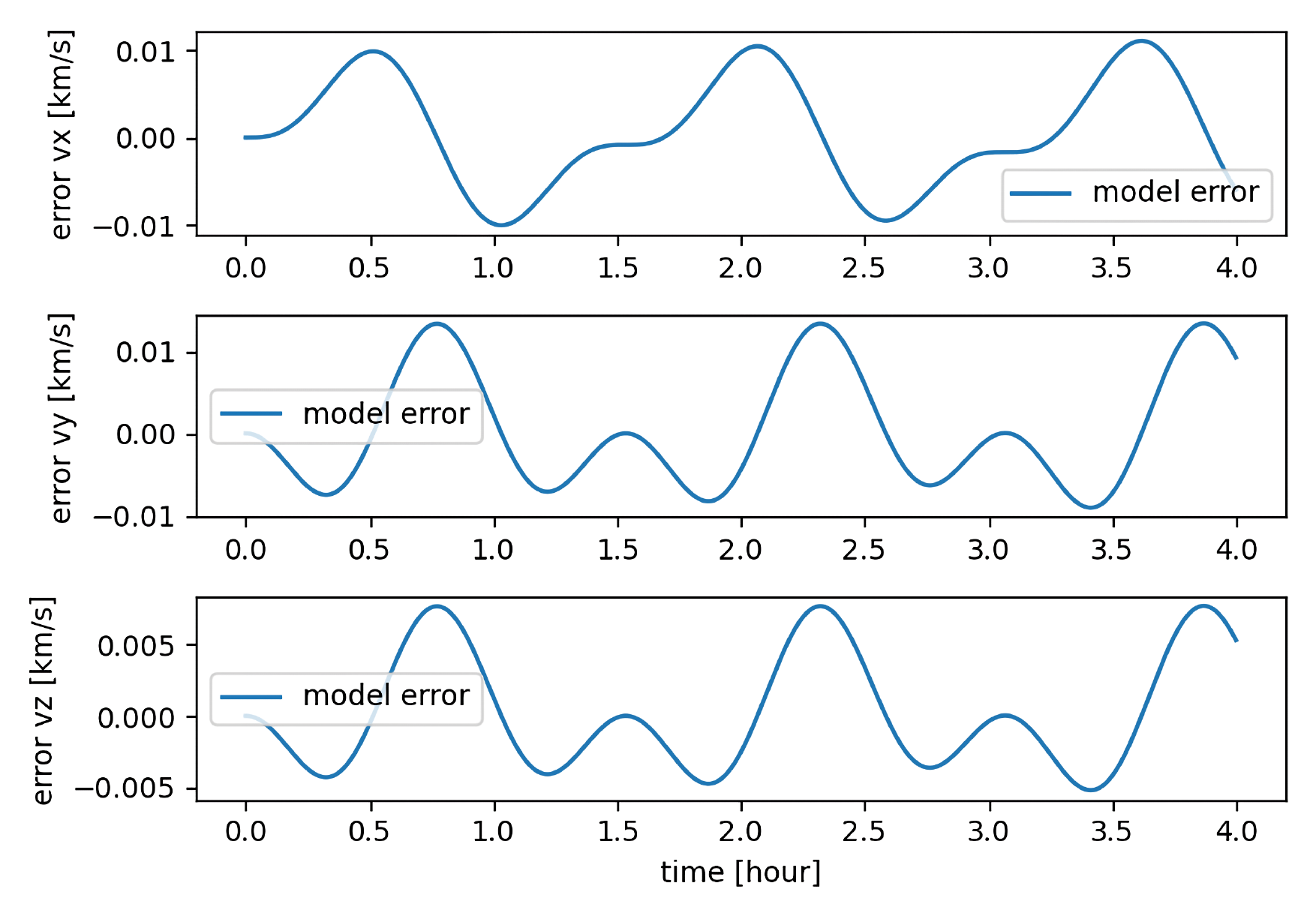}
        \caption{Errors in the velocity.}
        \label{fig:sub2}
    \end{subfigure}
    \hfill
    \begin{subfigure}[b]{0.3\linewidth}
        \includegraphics[width=\linewidth]{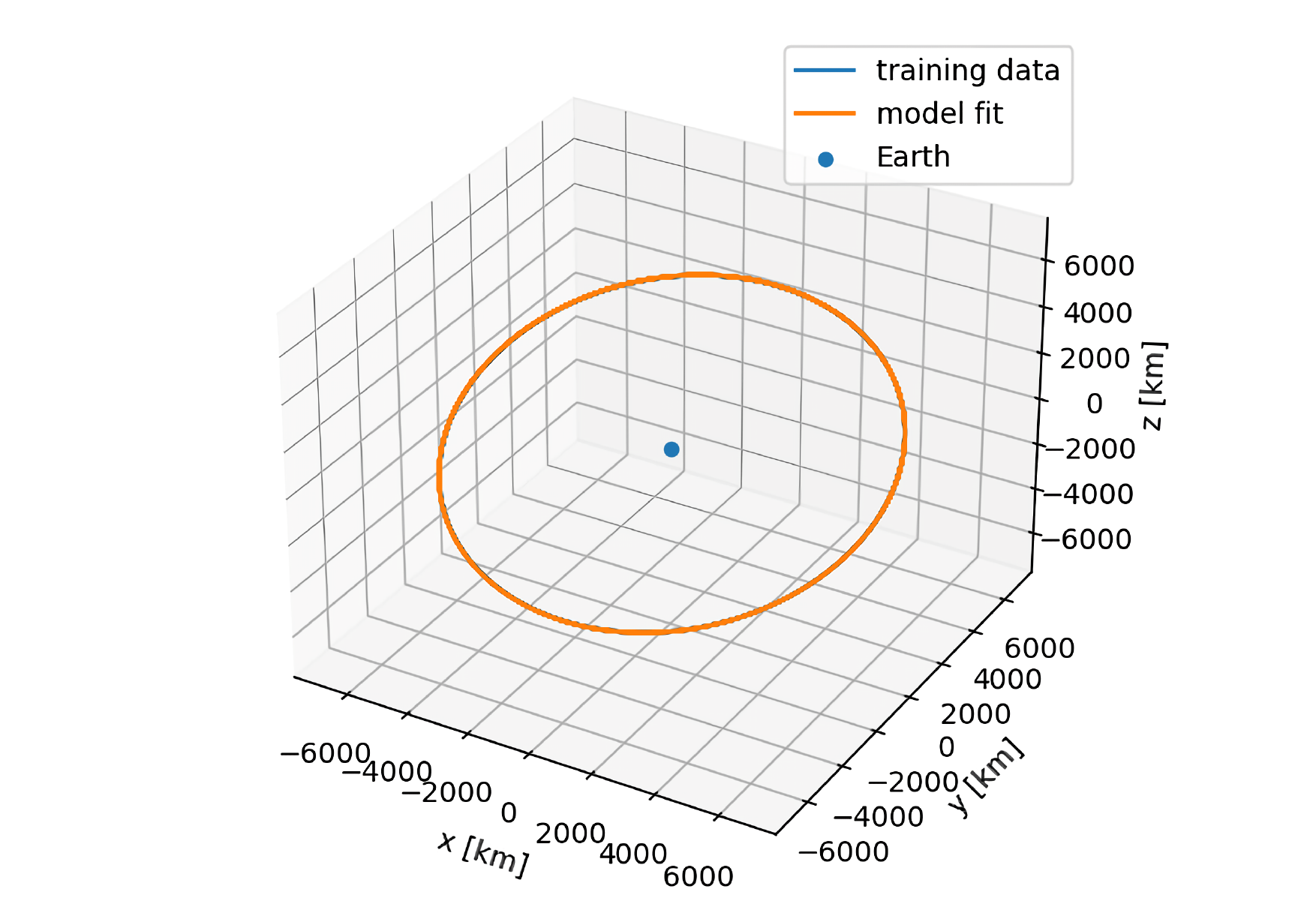}
        \caption{Orbital trajectory.}
        \label{fig:sub3}
    \end{subfigure}
    \caption[Single Orbit Performance]{Errors obtained when comparing the true values with the ones obtained using the polynomial library.}
    \label{fig:singorb}
\end{figure*}

When training the model with one orbit and simulating with it, the resultant errors on the positions reached a magnitude of around 10~km on the worst performing axis (i.e., the x-axis). On the velocities, the worst obtained error was approximately 0.01~km$s^{-1}$ as seen in Figure~\ref{fig:singorb}.

In our investigation, we also explored the utilisation of pre-computed derivatives in conjunction with SINDy. The idea was to supply the model with exact derivatives of the data points as an argument during the fitting process, referred to as using the $\Dot{x}$ argument. We pre-calculated the time derivatives based on the known differential equations governing the data, creating a matrix of these derivatives. However, upon careful evaluation, we observed that this approach yielded sub-optimal results compared to not using them, as it was done hitherto. Despite the initial appeal of mitigating noise amplification, the simulated orbit generated using pre-computed derivatives did not align with the expected trajectory. As a result, we have decided not to include this alternative approach in our final analysis, as it did not yield desirable outcomes.

\subsection{Multiple Trajectories} 
Another approach tested was learning from multiple satellite trajectories which supposedly would help the model have more data with a more complete spatial distribution. To do this, we selected the data from all the satellites and, for each one (for each ID), we integrated the differential equation so that we could get a dataset with no measurement noise (over a span of 3 hours). This resulted in a list with a length equal to the number of different satellites, where each index is a matrix representing the simulated dataset for that satellite. Along with this list of orbits, there is another list with the same length but with the corresponding times for each state vector of each satellite. Using this approach, we were getting worse results than only using one orbit, the errors were increasing, and the resulting orbit did not follow the correct trajectory. We found this counter-intuitive since learning with more data gave a worse performance.

To understand why adding more data to train the model resulted in worse predictions than only training with one orbit, we hypothesised that the problem could be that the orbits used were too similar and could be acting as noise rather than extra useful data. Consequently, instead of using each initial state propagated, the orbits were created by varying their inclinations, eccentricities, and altitudes to get a representative dataset of different trajectories. After generating this dataset and training the model with it, we applied it to see if it could accurately represent it given the initial state of an orbit. We created a different dataset to have more representative orbits with actual varying orbital elements as a way of us controlling what was being used for training. The orbits created had combinations of inclinations ranging from 0º to 180º with 15º intervals, altitudes from 200~km up to 1200~km with 200~km intervals, and three eccentricity values: 0.01, 0.02, and 0.03.

\begin{figure*}[t]
  \centering
  \begin{subfigure}[t]{0.3\linewidth}
    \centering
    \includegraphics[width=\linewidth]{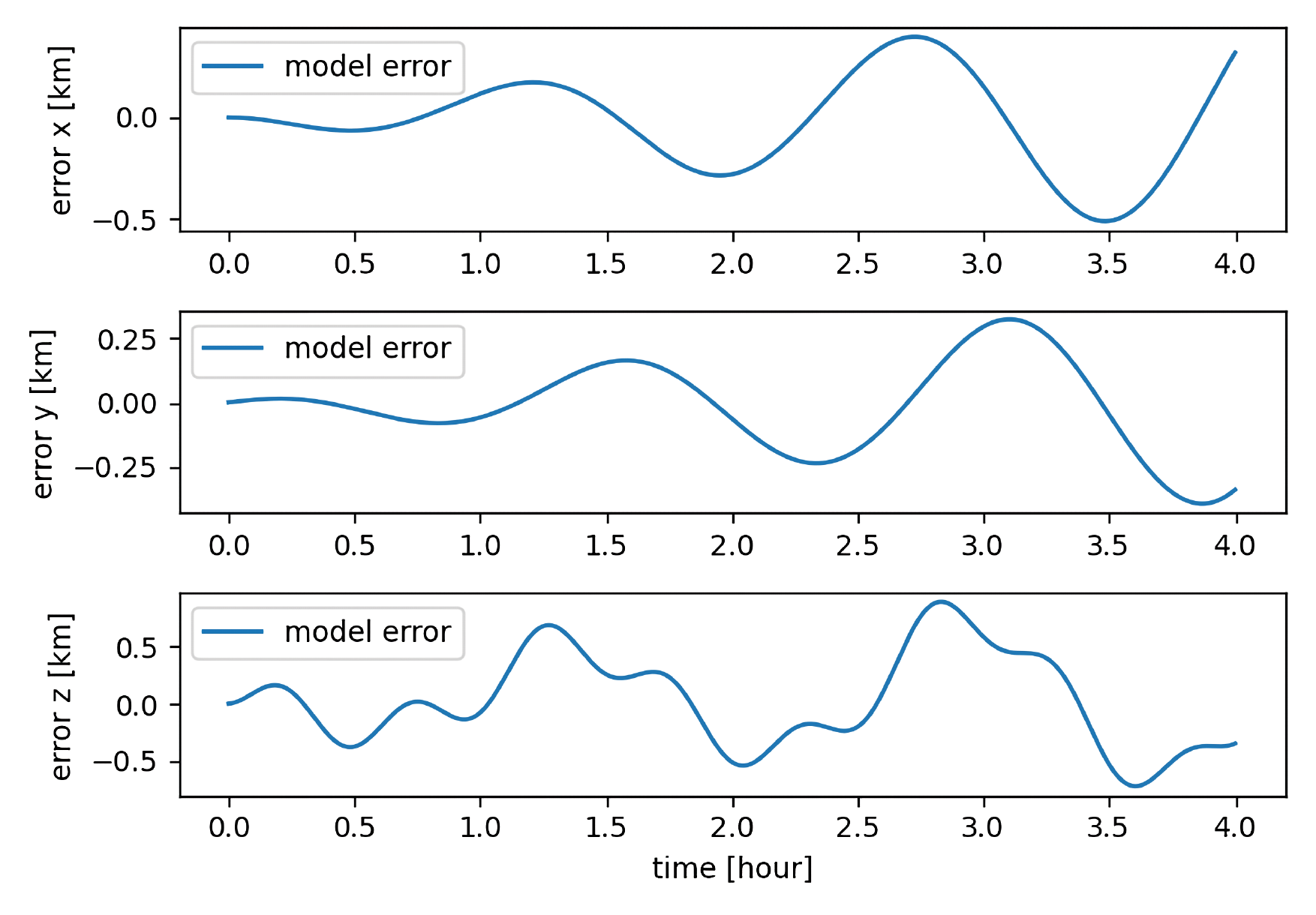}
    \caption{Errors in Positions}\label{fig:pos_err_multorbnodyn}
  \end{subfigure}%
  \hfill
  \begin{subfigure}[t]{0.3\linewidth}
    \centering
    \includegraphics[width=\linewidth]{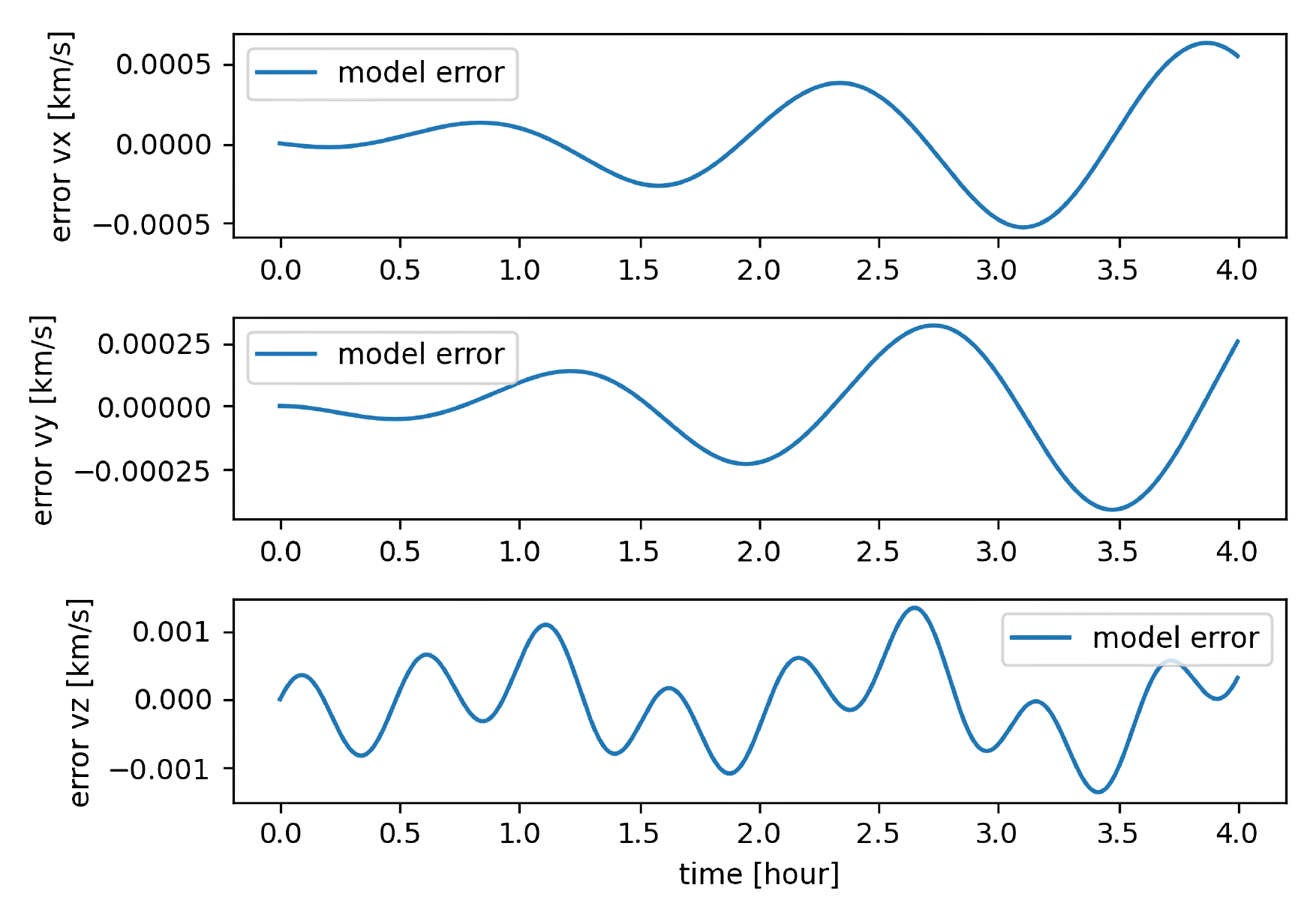}
    \caption{Errors in Velocities}\label{fig:vel_err_multorbnodyn}
  \end{subfigure}%
  \hfill
  \begin{subfigure}[t]{0.3\linewidth}
    \centering
    \includegraphics[width=\linewidth]{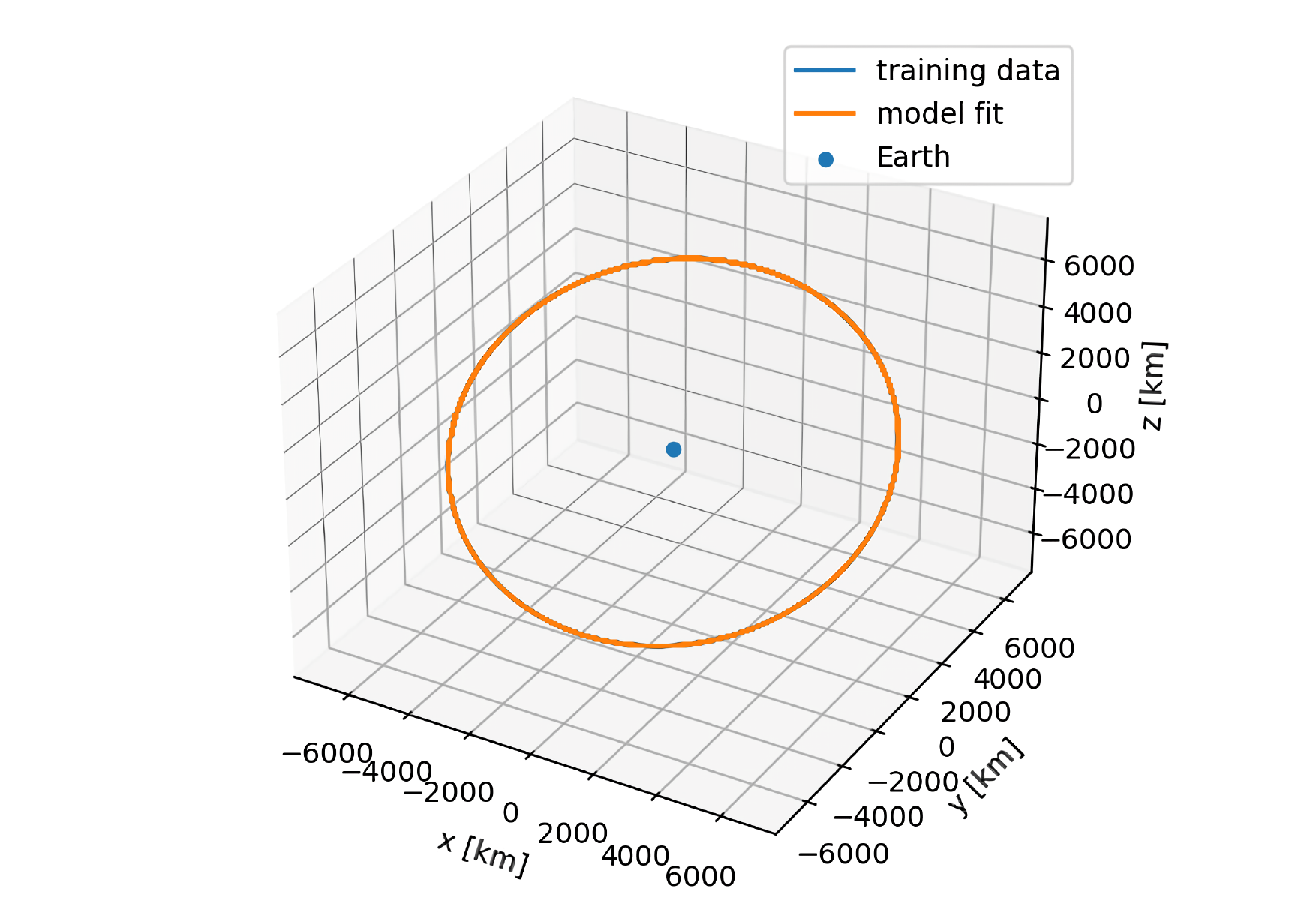}
    \caption{Orbit Dynamics}\label{fig:orbit_dyn_multorbnodyn}
  \end{subfigure}%
  \caption[Multiple Orbits Performance]{Graphics for the errors on the positions and the velocities and the resultant orbit learning with multiple trajectories}\label{fig:multorbnodyn}
\end{figure*}

When simulating the model with the same orbit used for the single orbit test but now training it with the multiple orbits aforementioned, the results improved drastically as seen in figure~\ref{fig:multorbnodyn}. This proves the hypothesis of the extra data not being representative. Using multiple orbits, it was possible to decrease the error on the positions on the worst performing axis to only around 0.5~km and for the velocities as low as 0.001~km$s^{-1}$. This is informative and proves beyond a reasonable doubt that learning with more representative data is better than with less data.

\section{Data with full dynamics}

Constructing a simulated dataset makes it possible to introduce realistic full dynamics related to drag and interactions with other celestial objects and see if~\gls{SINDy} manages to find the extra terms related to the added force. Drag corresponds to an extra term in the equations that define the derivative of the velocities which is subtracted from the acceleration value and is given by

$$\Vec{a}_{drag} = \frac{1}{2}\rho C_D \frac{A}{m} ||\Vec{v}|| \Vec{v}.$$

The dataset used corresponds to the same orbits with realistic values for $\rho$ resulting in realistic drag values and a more complex gravity modelling. It can be seen as a high-fidelity dataset with only differences from a real dataset due to noisy measurements.

\subsection{\textit{Custom Functions}}
Firstly, for testing, if it could find the exact terms for the equation that are known to represent drag, three extra functions had to be added to the library corresponding to the drag terms for each axis. Unfortunately, even with the presence of drag, making the satellite fall towards the earth at a rapid pace, the~\glspl{PDE} found, even though having extra terms compared to having no drag, the resulting orbit had the same contour as without drag which means the equations could not capture this effect. When repeating the same process for standardised data, a couple of residual terms appear on the equations, however, the orbit looks the same as without drag.

\subsection{\textit{Polynomial Functions}}
\begin{figure*}[t]
  \centering
  \begin{subfigure}[t]{0.3\linewidth}
    \includegraphics[width=\linewidth]{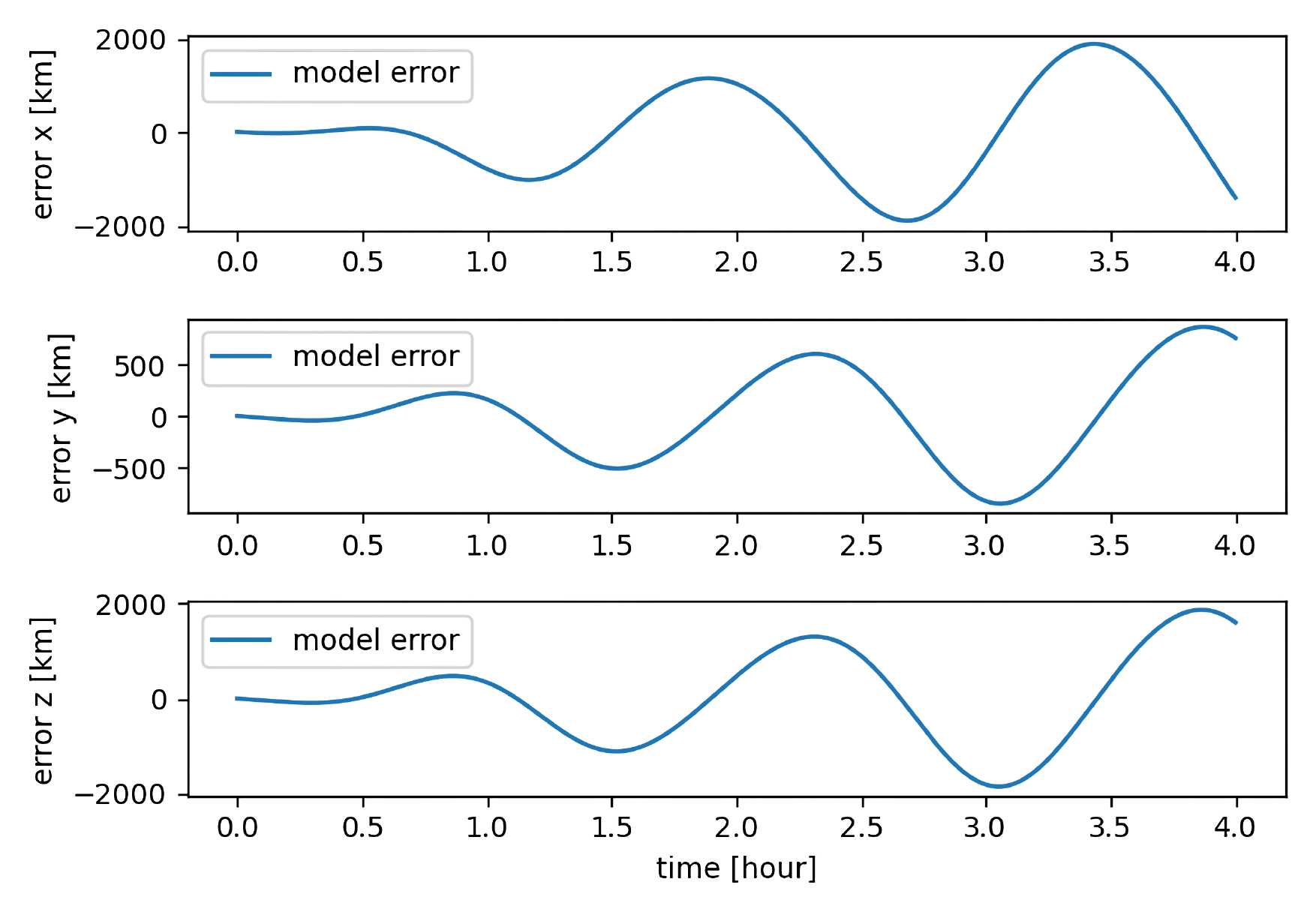}
    \caption{Errors in Positions}
    \label{fig:pos}
  \end{subfigure}
  \begin{subfigure}[t]{0.3\linewidth}
    \includegraphics[width=\linewidth]{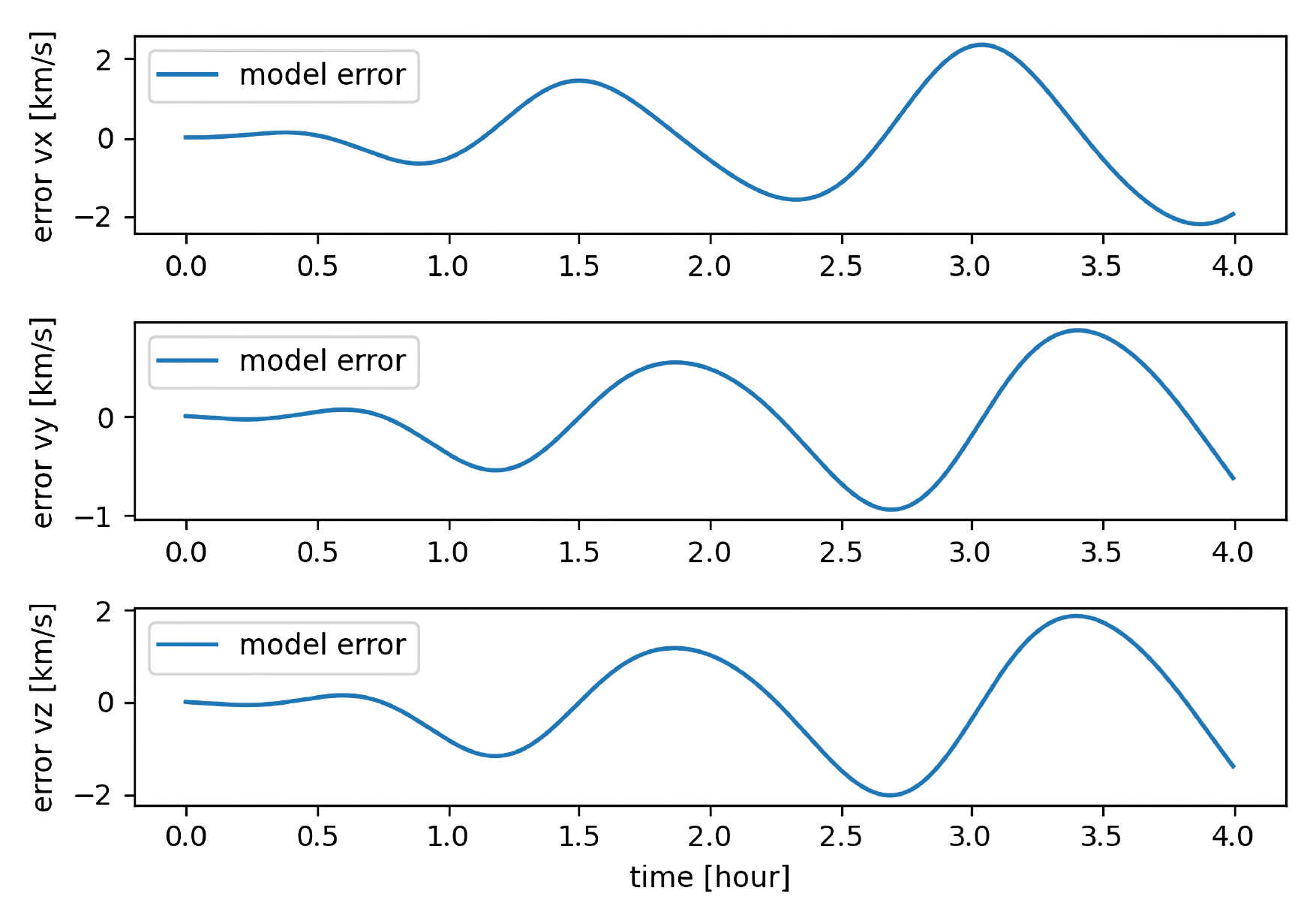}
    \caption{Errors in Velocities}
    \label{fig:vel}
  \end{subfigure}
  \begin{subfigure}[t]{0.3\linewidth}
    \includegraphics[width=\linewidth]{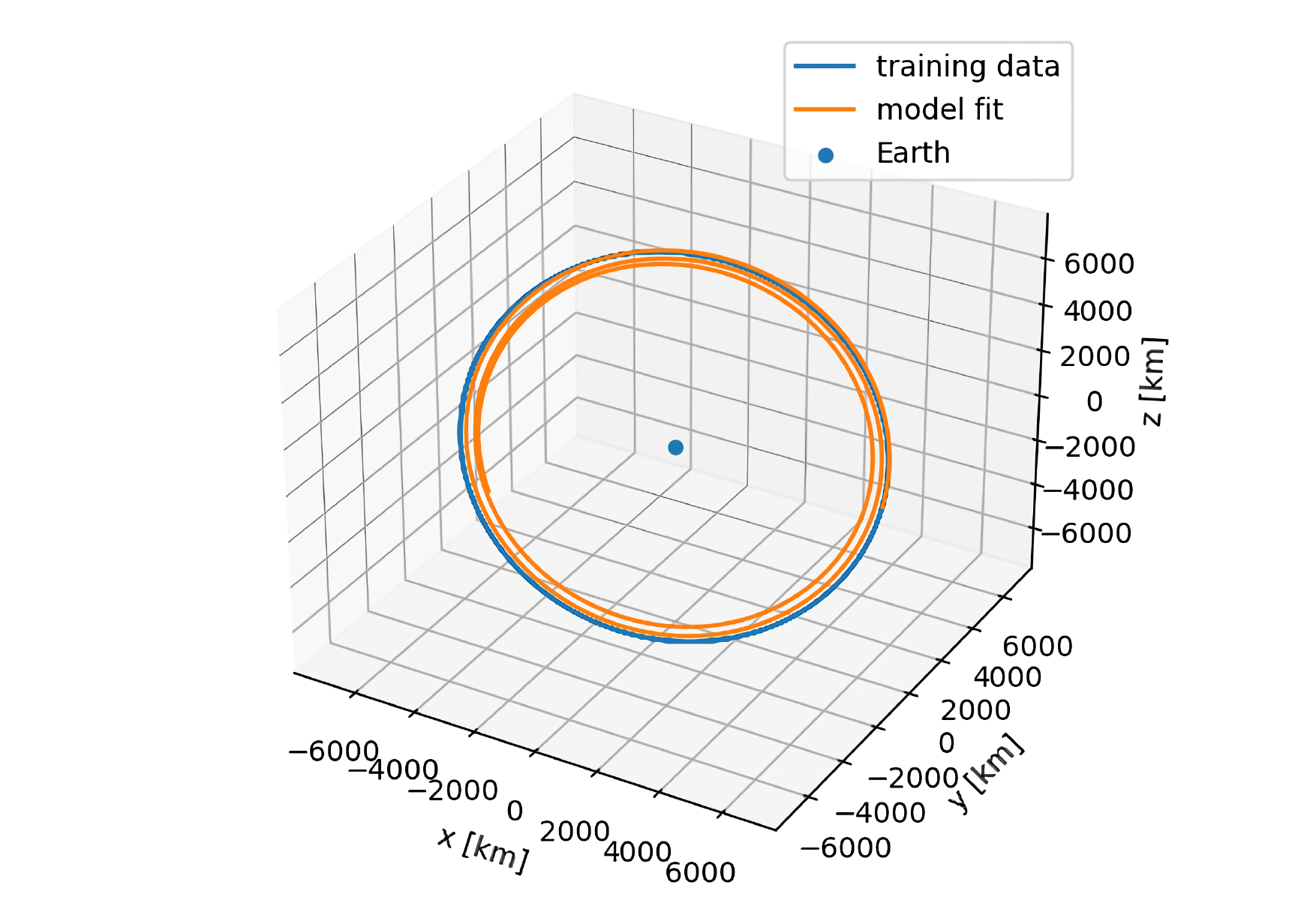}
    \caption{Orbit Dynamics}
    \label{fig:orbitdyn}
  \end{subfigure}
  \caption{Graphics for the errors on the positions and the velocities and the resultant orbit learning with one trajectory and full dynamics}
  \label{fig:singorbdyn}
\end{figure*}

Testing if the added polynomial terms could capture enhanced dynamics, repeating the process by training the model with only one orbit and simulating with a different one, the results were worse than simpler dynamics.
With standardised data, the orbit found spirals slightly into itself as seen in figure~\ref{fig:singorbdyn} which might be an amplification of what happens due to the decrease in the altitude of the satellite.

\subsection{\textit{Multiple Trajectories}}
\label{section:multtraj}
\begin{figure*}[t]
  \centering
  \subfloat[Errors in Positions\label{fig:rightsubfig1}]{%
    \includegraphics[width=0.3\linewidth]{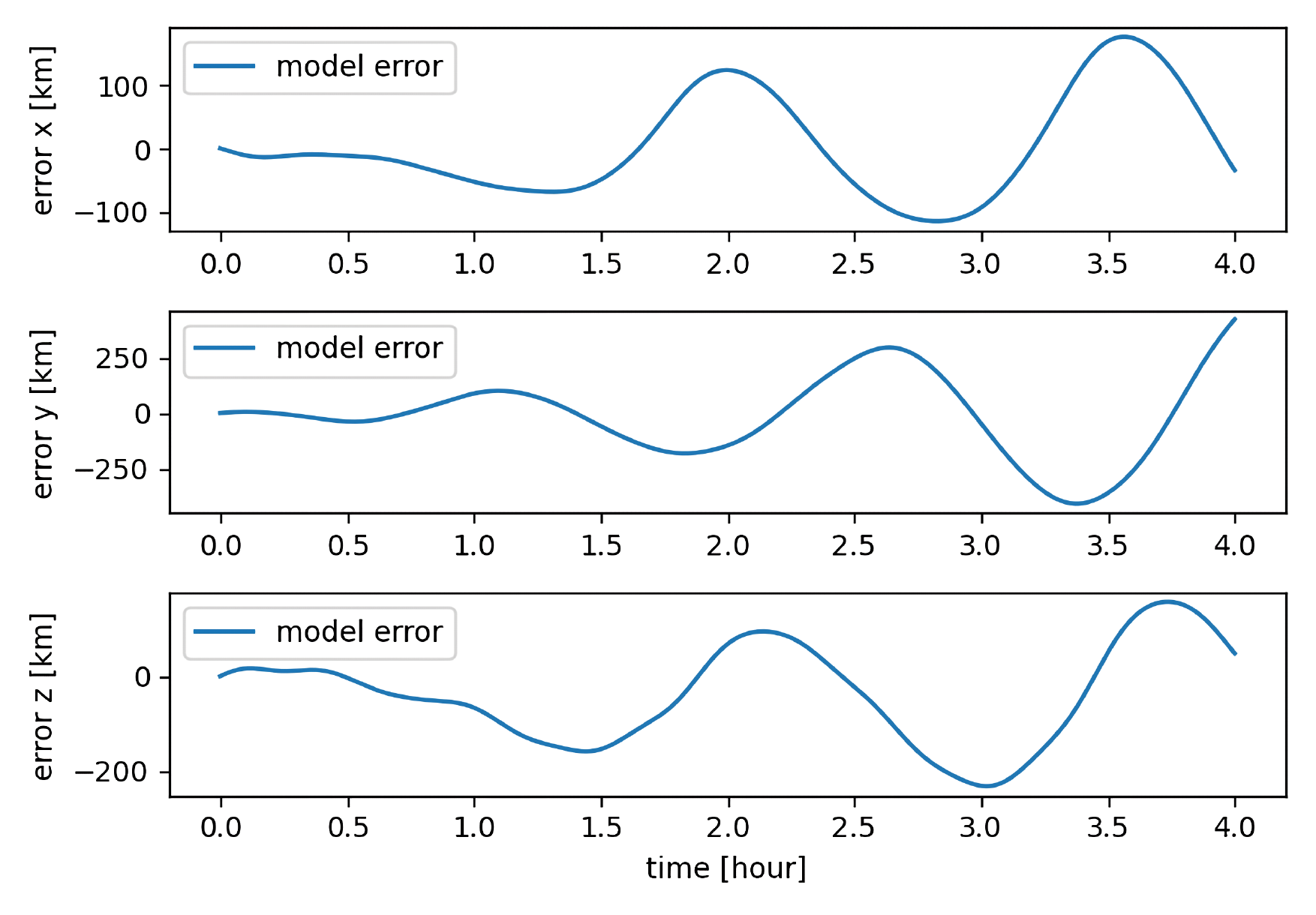}}
   \hfill
   \subfloat[Errors in Velocities\label{fig:rightsubfig2}]{%
    \includegraphics[width=0.3\linewidth]{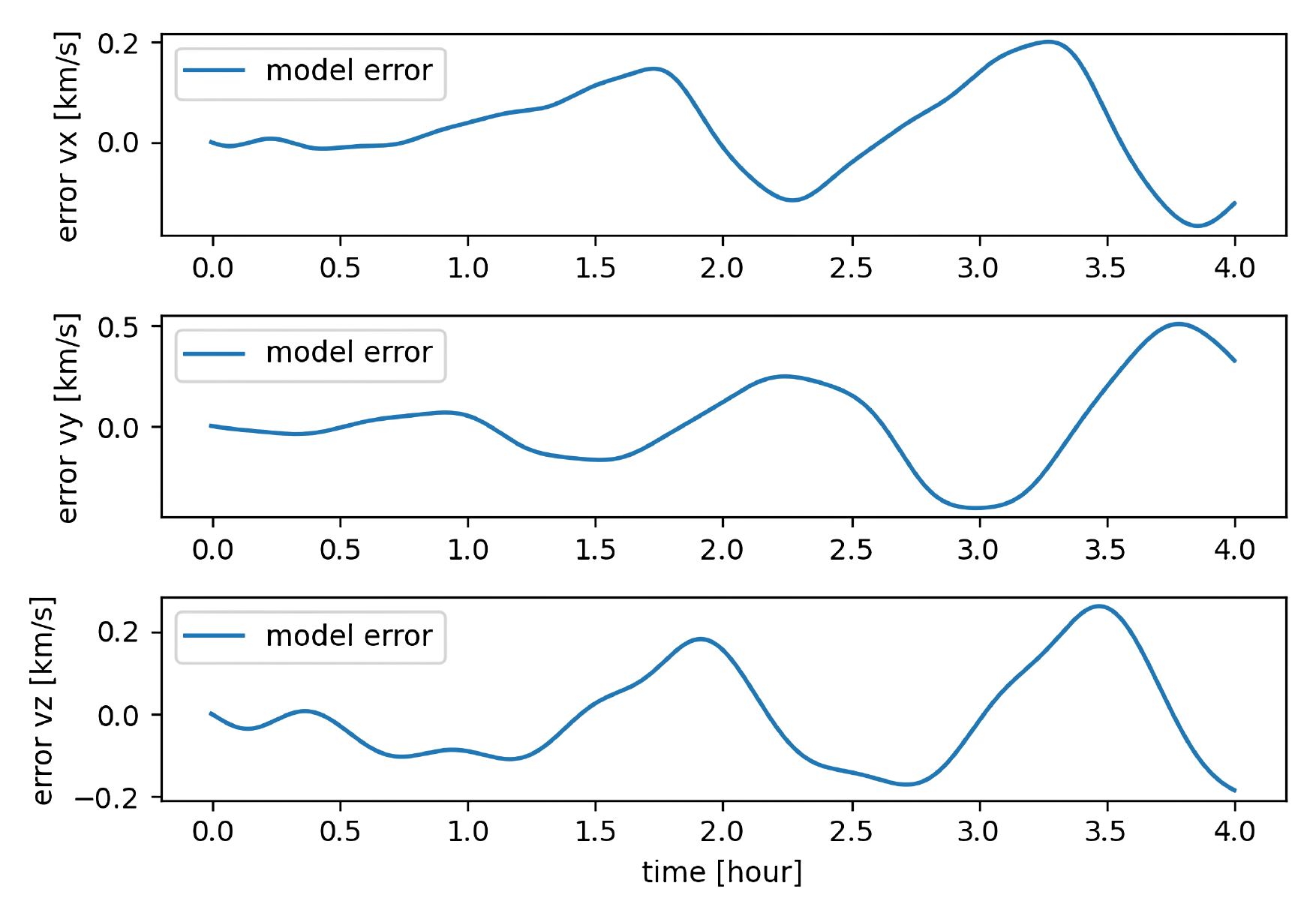}}
    \hfill
    \subfloat[Orbit Dynamics\label{fig:rightsubfig3}]{%
    \includegraphics[width=0.3\linewidth]{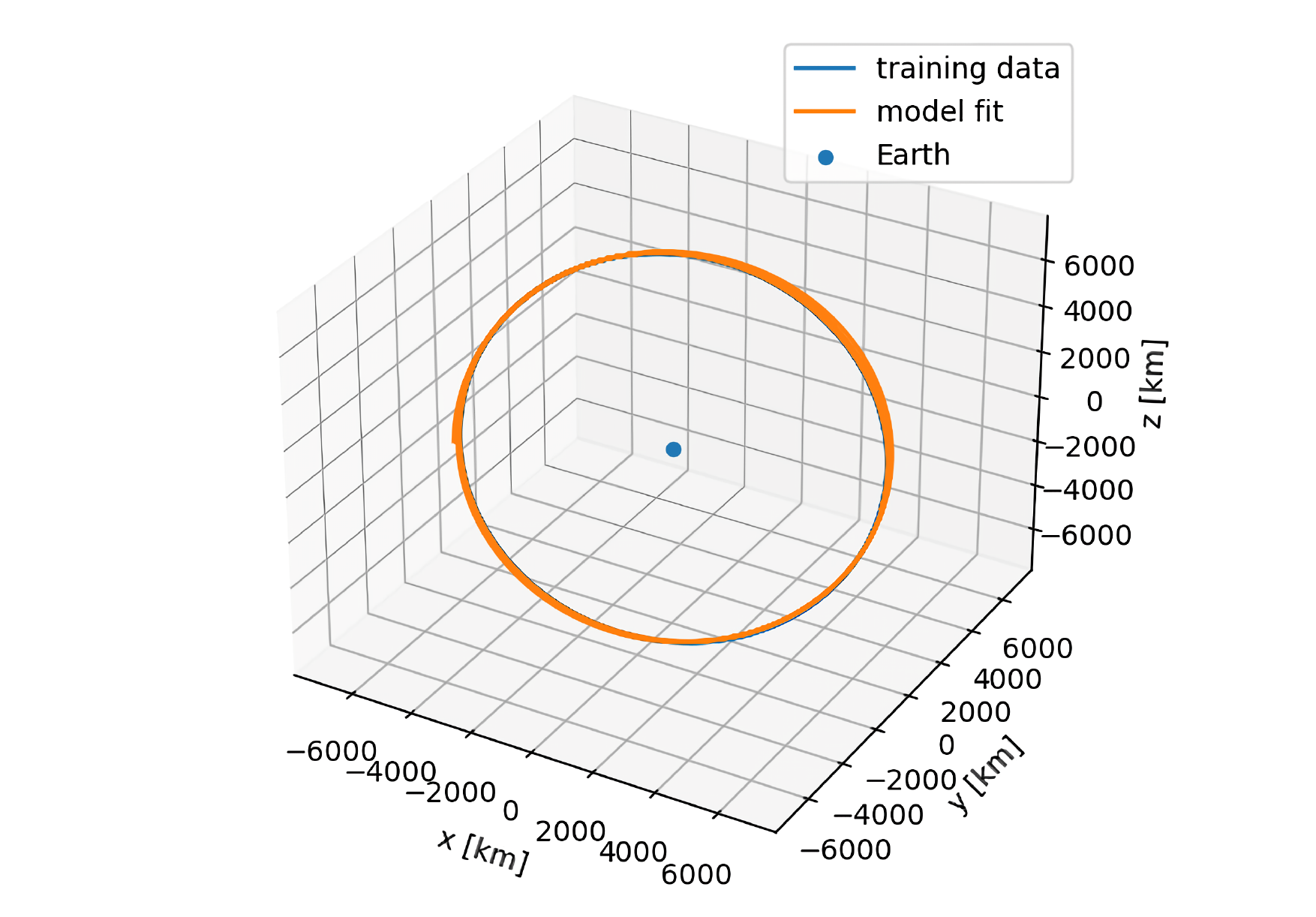}}
  \caption[Multiple Orbits Performance with full dynamics]{Graphics for the errors on the positions and the velocities and the resultant orbit learning with multiple trajectories and full dynamics}
  \label{fig:multorbdyn}
\end{figure*}

Using standardised data and training the model with the descriptive high-fidelity dataset, the results were much better. Training with multiple trajectories and simulating with a different one with an unseen inclination, eccentricity, and altitude, decreased the error compared to only using a realistic orbit. In this case, it is possible to see the outline of the orbit in Figure~\ref{fig:multorbdyn} follows much more precisely the correct trajectory and does not suffer from incorrect spiralling.

\section{Noise Analysis: Robustness to heavy-tailed noise}

\begin{table}[!htb]
\renewcommand{\arraystretch}{1.3}
\caption[Noise Analysis - Custom Library]{Robustness of SINDy to multiple types of noise - Custom Library.}
\label{table:custom}
\centering
\begin{tabular}{ccc}
\hline
\textbf{Noise Type} & \textbf{Position} & \textbf{Velocity} \\
\hline
Gaussian Noise     & 0.1\%    & 1\%      \\
Laplacian Noise    & 0.1\%    & 0.1\%    \\
Cauchy Noise       & 0.001\%  & 0.01\%   \\
\hline
\end{tabular}
\end{table}

\begin{table}[!htb]
\renewcommand{\arraystretch}{1.3}
\caption[Noise Analysis - Polynomial Library]{Robustness of SINDy to multiple types of noise - Polynomial Library.}
\label{table:poly}
\centering
\begin{tabular}{ccc}
\hline
\textbf{Noise Type} & \textbf{Position} & \textbf{Velocity} \\ \hline
Gaussian Noise     & 0.37\%   & 0.79\%   \\
Laplacian Noise    & 0.38\%   & 0.53\%   \\
Cauchy Noise       & 0.01\%   & 0.1\%    \\
\hline
\end{tabular}
\end{table}

\begin{figure*}[htbp]
  \centering
  \includegraphics[width=0.6\textwidth]{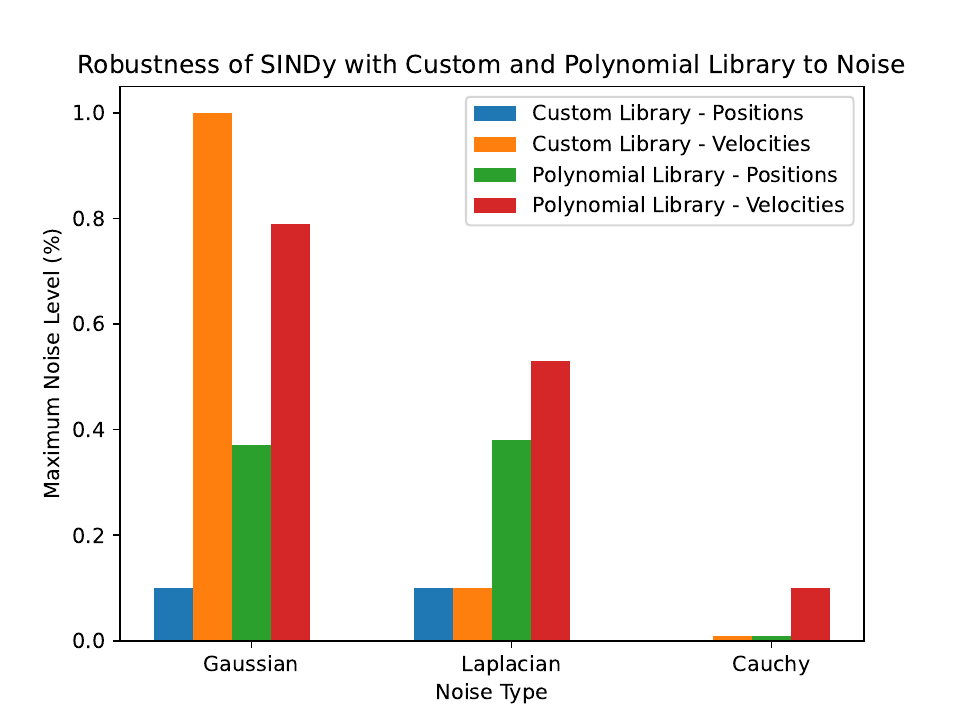}
  \caption{Robustness of SINDy to Different Types of Noise}
  \label{fig:robustness_plot}
\end{figure*}

To see how capable~\gls{SINDy} is in discovering~\glspl{PDE} in multiple noise conditions, considering that its performance excelled in discovering first-order~\glspl{PDE}, we introduced multiple types of noise to the data and then used~\gls{SINDy} to discover those first-order equations to check how robust it is, given a noisy input using the custom library function and the polynomial library whose results are shown respectively in Table \ref{table:custom} and \ref{table:poly}. There were three types of noise we tested~\gls{SINDy} with due to their intrinsic characteristics. The first one we tested with was the Gaussian noise. One of the biggest advantages is that when dealing with a lot of data, it tends to respect the central limit theorem that the Gaussian distribution describes. Having thin tails and the majority of its probability mass around the mean also represents having a lower probability of generating outliers. 
Afterwards, we applied two other different types of noise: Laplacian and Cauchy. The distributions that describe these two types of noise are heavy-tailed, so they are usually used for modelling outliers as they appear in the distribution areas further away from the mean. Laplacian noise can be considered two exponential distributions pointy around the mean whereas the Cauchy distribution has even heavier tails to the point where a mean value does not even parameterise it. One important step to remember is that the noise may have a different effect depending on which variables are affected, whether the noise is present in the positions or the velocities. We calculated the norm of a position vector and applied the different types of noise with a standard deviation varying from 0 to 10\% of the norm of that vector and did the same for the velocities. The noise robustness associated with both the custom and polynomial libraries, encompassing the three distinct noise types and their impact on the positions and velocities, are depicted in Figure~\ref{fig:robustness_plot}.

Using the custom library, we utilised~\gls{SINDy} to discover first-order equations in the presence of noise. For \textbf{Gaussian noise}, if there is only noise on the velocity,~\gls{SINDy} supports up to 1\% noise before yielding non-explanatory equations for the system. On the other hand, if there is only noise on the position,~\gls{SINDy} supports up to 0.1\% of noise. Regarding 1\% noise on the velocity,~\gls{SINDy} does not support any noise on the position. However, with 0.1\% noise on the velocity,~\gls{SINDy} supports as low as $10^{-6}$\% noise on the position.

Regarding Laplacian noise, \gls{SINDy} demonstrates similar robustness in handling noise in the positions, but this ability decreases with velocities. If there is only noise on the velocities, \gls{SINDy} allows up to 0.1\% noise compared to the 1\% tolerance with Gaussian noise. Similarly, if there is only noise on the positions, \gls{SINDy} allows up to 0.1\% noise, consistent with the Gaussian noise scenario.

Concerning Cauchy noise, it has two parameters, \textit{loc} which specifies where the peak of the distribution will be on the X axis (which is 0 by default) and a scale parameter representing half the width of the PDF at half the maximum height. Fixing the position parameter and varying the scaling factor, having only noise on the positions,~\gls{SINDy} supports noise up to 0.001\% of the position vector norm. For velocities, it supports noise up to 0.01\% of the norm of a velocity vector, which is 1 order of magnitude less.

Considering the polynomial library, which has shown promising results, evaluating its performance in the presence of noise is crucial. The analysis in this section focuses on standardised data. For Gaussian noise, if there is only noise on the positions,~\gls{SINDy} supports a standard deviation up to 0.37\% of the norm of the first position vector, corresponding to a noise level of up to 26 km in positions, which is reasonable. Regarding velocities,~\gls{SINDy} supports up to 0.79\%, equivalent to 0.06 km/s. When the standard deviation for the positions is 26~km, the same percentage of noise is supported for velocities (0.79\%).

Results for Laplacian noise are similar, with support for up to 27~km noise in positions and up to 0.04~km/s noise in velocities when noise is present only in the respective variables. In the case of maximum supported position noise (27~km), the supported noise in velocities remains at 0.04~km/s. 

Lastly, for Cauchy noise,~\gls{SINDy} supports up to 0.01\% of the norm of the positions and 0.1\% of the norm of the velocities.

Comparing the three types of noise, it is evident that~\gls{SINDy} demonstrates greater robustness to noise on the velocities than on the positions. Additionally, it is important to note that when using the polynomial library with non-standardised data, \gls{SINDy} struggles to identify any~\glspl{PDE} for the velocities, regardless of the introduced noise level. This emphasises the importance of standardising the data before applying \gls{SINDy} to achieve more reliable results.

\section{Future Work}
As this work demonstrated, the usage of~\gls{SINDy} offers a promising approach for discovering the equations that govern the behaviour of satellites in space. However, there is still significant room for accuracy and predictive power improvement. In the future, one potential avenue for further improving the results obtained in this study is to incorporate the found equations into the architecture of a ~\gls{PINN}~\cite{pinnsProposed}.

Using the found equations as soft constraints within the~\gls{PINN} may further refine the predictions of the behaviour of the system under different conditions. This approach has already shown promise in other domains, and there is reason to believe it could also be successful in satellite dynamics. Further exploration of different function libraries and optimisation techniques may lead to even better results. 

\section{Conclusion}
This work has demonstrated the effectiveness of the~\gls{SINDy} algorithm in identifying the equations that govern the behaviour of satellites in space. The results obtained using \gls{SINDy} were highly accurate and showed excellent predictive power, highlighting the potential of this approach for improving our understanding of space systems.

In particular, it allowed us to identify the underlying dynamics of the satellite system, which is critical for developing accurate models and predictions of the behaviour of the system. Furthermore, comparing the propagation of the identified equations with the actual state vectors for the future of those satellites revealed a high degree of accuracy, demonstrating the efficacy of the \gls{SINDy} algorithm.

Overall, the results obtained in this study offer valuable insights into the behaviour of space systems and provide a promising foundation for future research in this area. With the continued development of \gls{SINDy} and other machine learning techniques, we will likely see further improvements in our ability to model and predict the behaviour of satellites or debris, ultimately contributing to better tracking of those objects.

\section*{Acknowledgements}
This research was carried out under Project “Artificial Intelligence Fights Space Debris” Nº C626449889-0046305 co-funded by Recovery and Resilience Plan and NextGeneration EU Funds, \url{www.recuperarportugal.gov.pt}.

\bibliographystyle{unsrtnat}
% \bibliography{references}

%%%%%%%%%%%%%%%%%%%%%%%%%%%%%%%%%%%%%%%%%%%%%%%%%%%%%%%%%%%%%%%%%%%%%%

\end{document}